\documentclass[12pt, a4paper]{article}
\usepackage[numbers,sort&compress]{natbib}
\usepackage[final]{nips_2017}
\usepackage[utf8]{inputenc}
\usepackage[T1]{fontenc}
\usepackage{url, bbm}
\usepackage{hyperref}
\usepackage{amsmath}
\usepackage{amsfonts}
\usepackage{pifont}
\usepackage{amssymb}
\usepackage{nicefrac}
\usepackage{mathtools}
\usepackage{cleveref}
\usepackage{enumerate}
\usepackage{xifthen}
\usepackage{xargs}
\usepackage{wrapfig}
\usepackage{placeins}
\usepackage{stix}
\usepackage{algorithm,algpseudocode}
\usepackage{enumitem}
\usepackage[titletoc,title]{appendix}
\usepackage{caption}
\usepackage{tabularx}

\newcommand{\st}{\text{s.t.}}
\newcommand{\wrt}{\text{w.r.t.}}

\newcommand{\perse}{\emph{per se}}

\newcommand{\alt}[1]{\ensuremath{\tilde{#1}}}
\newcommand{\transform}[1]{\ensuremath{#1^{\prime}}}

\newcommand{\transpose}[1]{\ensuremath{#1^\top}}

\newcommand{\nsamples}{\ensuremath{n}}
\newcommand{\batchsize}{\ensuremath{m_{b}}}

\newcommand{\x}{\ensuremath{\mathbf{x}}}
\newcommand{\X}{\ensuremath{\mathbf{X}}}

\newcommand{\y}{\ensuremath{y}}
\newcommand{\Y}{\ensuremath{\mathbf{y}}}
\newcommand{\yspace}{\ensuremath{\mathcal{Y}}}

\newcommand{\z}{\ensuremath{z}}
\newcommand{\Z}{\ensuremath{\mathbf{z}}}
\newcommand{\zspace}{\ensuremath{\mathcal{Z}}}

\newcommand{\realSymb}{\mathbb{R}}
\newcommand{\real}[1]{\ensuremath{\ifthenelse{\isempty{#1}{}}{\realSymb}{\realSymb^{#1}}}}

\newcommand{\zeros}{\ensuremath{\mathbf{0}}}
\newcommand{\identity}{\ensuremath{\mathbf{I}}}

\newcommand{\expectation}[1]{\ensuremath{\mathbb{E}\left[#1\right]}}
\newcommand{\abs}[1]{\ensuremath{\lvert#1\rvert}}

\newcommand{\gaussian}{\ensuremath{\mathcal{N}}}

\newcommand{\means}{\ensuremath{\boldsymbol{\mu}}}
\newcommand{\mean}{\ensuremath{\mu}}

\newcommand{\Cov}{\ensuremath{\boldsymbol{\Sigma}}}

\newcommand{\chol}{\ensuremath{L}}
\newcommand{\Chol}{\ensuremath{\mathbf{L}}}
\newcommand{\stdv}{\ensuremath{\sigma}}

\newcommand{\var}{\ensuremath{\stdv^{2}}}

\newcommand{\region}{\ensuremath{\mathcal{A}}}

\newcommand{\poolsize}{\ensuremath{q}}
\newcommand{\temperature}{\ensuremath{\tau}}

\newcommand{\f}{\ensuremath{f}}

\newcommand{\g}{\ensuremath{h}}
\newcommand{\G}{\ensuremath{H}}

\newcommand{\rparams}{\ensuremath{\boldsymbol{\theta}}}

\newcommand{\gparams}{\ensuremath{\boldsymbol{\phi}}}

\newcommand{\reparam}{\ensuremath{\rho}}

\newcommand{\lbounds}{\ensuremath{\mathbf{a}}}

\newcommand{\ubounds}{\ensuremath{\mathbf{b}}}

\newcommand{\sigmoid}{\ensuremath{\operatorname{\sigma}}}
\newcommand{\softmax}{\ensuremath{\operatorname{softmax}}}
\newcommand{\heaviside}{\ensuremath{\operatorname{\mathbbm{1}}}}

\newcommand{\mathhyphen}{{\hbox{-}}}
\newcommand{\EI}{\ensuremath{\operatorname{EI}}}
\newcommand{\qEI}{\ensuremath{\operatorname{\mathnormal{q}\mathhyphen{}EI}}}
\newcommand{\UCB}{\ensuremath{\operatorname{UCB}}}
\newcommand{\qUCB}{\ensuremath{\operatorname{\mathnormal{q}\mathhyphen{}UCB}}}
\newcommand{\mUCB}{\ensuremath{\operatorname{\mathnormal{1}\mathhyphen{}UCB}}}
\newcommand{\PI}{\ensuremath{\operatorname{PI}}}
\newcommand{\qPI}{\ensuremath{\operatorname{\mathnormal{q}\mathhyphen{}PI}}}

\newcommand{\qSR}{\ensuremath{\operatorname{\mathnormal{q}\mathhyphen{}SR}}}
\newcommand{\ES}{\ensuremath{\operatorname{ES}}}
\newcommand{\qES}{\ensuremath{\operatorname{\mathnormal{q}\mathhyphen{}ES}}}

\newcommand{\rmsprop}{\text{RMSProp}}
\newcommand{\adam}{\text{Adam}}
\newcommand{\direct}{\text{DIRECT}}

\newcommand{\model}{\ensuremath{\mathcal{M}}}
\newcommand{\data}{\ensuremath{\mathcal{D}}}

\DeclarePairedDelimiter{\floor}{\lfloor}{\floor}

\algnewcommand\blank[1]{\item[#1]}
\algnewcommand\cmd[1]{\item[\textbf{#1:}]}
\algnewcommand\while[1]{\item[\textbf{while} ({\small#1})]}
\algrenewcommand\algorithmicindent{.8em}

\usepackage[pdftex,dvipsnames]{xcolor}
\usepackage[colorinlistoftodos,prependcaption,textsize=tiny]{todonotes}

\definecolor{mgray}{gray}{0.5}
\definecolor{dblue}{rgb}{0.0,0.0,0.5}
\definecolor{dred}{rgb}{0.5, 0.0, 0.0}

\newcommandx{\flag}[2][1=]
  {\todo[inline,linecolor=red,backgroundcolor=red!25,bordercolor=red,#1]{#2}}
\newcommandx{\ask}[2][1=]
  {\todo[inline,linecolor=OliveGreen,backgroundcolor=OliveGreen!25,bordercolor=OliveGreen,#1]{#2}}
\newcommandx{\remind}[2][1=]
  {\todo[inline,linecolor=Gray,backgroundcolor=Gray!25,bordercolor=Gray,#1]{#2}}
\newcommandx{\note}[2][1=]
  {\todo[inline,linecolor=blue,backgroundcolor=blue!25,bordercolor=blue,#1]{#2}}

\title{The reparameterization trick for acquisition functions}
\author{
  James T. Wilson\textsuperscript{1,2}\\
  \texttt{j.wilson17@imperial.ac.uk}\\
  \And
  Riccardo Moriconi\textsuperscript{1}\\
  \texttt{r.moriconi16@imperial.ac.uk}\\
  \And
  Frank Hutter\textsuperscript{2}\\
  \texttt{fh@cs.uni-freiburg.de}\\
  \And
  Marc Peter Deisenroth\textsuperscript{1}\\
  \texttt{m.deisenroth@imperial.ac.uk}\\
  \and
  \textsuperscript{1}\footnotesize{Imperial College of London}
  \hspace{1.85cm}
  \and
  \textsuperscript{2}\footnotesize{University of Freiburg}
  \hspace{0.65cm}
}

\begin{document}
\maketitle

\begin{abstract}
\label{sect:abstract}
Bayesian optimization is a sample-efficient approach to solving global optimization problems. Along with a surrogate model, this approach relies on theoretically motivated value heuristics (acquisition functions) to guide the search process. Maximizing acquisition functions yields the best performance; unfortunately, this ideal is difficult to achieve since optimizing acquisition functions \perse{} is frequently non-trivial. This statement is especially true in the parallel setting, where acquisition functions are routinely non-convex, high-dimensional, and intractable. Here, we demonstrate how many popular acquisition functions can be formulated as Gaussian integrals amenable to the reparameterization trick \cite{JimenezRezende2014a, kingma2013} and, ensuingly, gradient-based optimization. Further, we use this reparameterized representation to derive an efficient Monte Carlo estimator for the upper confidence bound acquisition function \cite{srinivas10} in the context of parallel selection.
\end{abstract}

\section{Introduction}
\label{sect:introduction}
In Bayesian optimization (BO), acquisition functions \G{}, with few exceptions, amount to integrals defined in terms of a belief \(p\) over the unknown outcomes \(\Y = \{\y_{1},\ldots,\y_{\poolsize}\}\) revealed when evaluating a black-box function \f{} at corresponding input locations \(\X = \{\x_{1},\ldots,\x_{\poolsize}\}\). This formulation naturally occurs as part of a Bayesian approach whereby we would like to assess how valuable different queries \X{} are to the optimization process by accounting for all conceivable realizations of \(\Y = \f(\X)\). Denoting by \g{} the function used to convey the value-added for observing a given realization, this paradigm gives rise to acquisition functions defined as
\begin{align}
\label{eq:acquisition_integral}
\G(\X; \gparams) &= \int_{\region} \g(\Y; \gparams) p(\Y|\X, \data{}) d\Y\,,
\end{align}
where integration region \(\region \subseteq \yspace\) represents the set of all possible outcomes \Y, \gparams{} any additional parameters associated with integrand \g{}, and \data{} the available prior information.\footnote{Henceforth, we omit explicit reference to prior information \data{}.} Without loss of generality, we express acquisition functions as \(\poolsize\)-dimensional integrals, where \(\poolsize\) denotes the total number of queries with unknown outcomes after each decision. For pool-size \(\poolsize = 1\), we recover strictly sequential decision-making rules; whereas, for \(\poolsize > 1\), we obtain strategies for parallel selection.\footnote{To avoid confusion when discussing SGD, we reserve the term \emph{batch-size} for description of minibatches.} As an exception to this rule, \emph{non-myopic} acquisition functions, which assign value by further considering how different realizations of \((\X,\Y)\) impact our broader understanding of black-box \f{}, generally correspond to higher-dimensional integrals. Specifically, non-myopic instances of the above formulation typically recurse, with the integrand \g{} amounting to an additional integral of the form~\eqref{eq:acquisition_integral}. While in a minority of cases closed-form solutions exist, these integrals are generally intractable and therefore difficult to optimize.

For this reason, a variety of methods have been proposed for evaluating intractable acquisition functions. These approaches have ranged from expectation propagation-based approximations of Gaussian probabilities~\cite{cunningham2011epmgp, hennig-jmlr12a, hernandez-nips14} to bespoke approximation strategies~\cite{contal2013parallel,desautels2014parallelizing} to sample-based Monte Carlo techniques~\cite{osborne2009gaussian,snoek-nips12a,hennig-jmlr12a}.

The special case of parallel Expected Improvement (\qEI{}) has received considerable attention \cite{ginsbourger2010kriging,snoek-nips12a,chevalier2013fast,wang2016parallel}; however, excepting \cite{wang2016parallel}, proposed methods do not scale gracefully in pool-size \poolsize{}. Still within the context of \qEI{} and independent of our work, \cite{wang2016parallel} derive results analogous to our own, but refer to the reparameterization trick (discussed below) as \emph{infinitesimal perturbation analysis} \cite{glasserman2013monte}.

In this work, we focus on the most common estimation technique: Monte Carlo integration. Despite their generality and myriad other desirable properties, Monte Carlo approaches have consistently been regarded as non-differentiable and, therefore, inefficient in practice given the need to optimize~\eqref{eq:acquisition_integral}. However, it seems to have been overlooked that sample-based approaches can indeed be used to estimate gradients, well-known examples of which include stochastic backpropagation and the reparameterization trick \cite{JimenezRezende2014a,kingma2013}. In the following, we exploit this insight to demonstrate gradient-based optimization of acquisition functions estimated via Monte Carlo integration.

The \emph{reparameterization trick} is a way of rewriting functions of random variables that makes their differentiability \wrt{} the parameters of an underlying distribution transparent. The trick applies a deterministic mapping \(\reparam : \zspace \to \yspace\) from random variables \(\Z \in \zspace\) with a parameter-free base distribution to random variables \(\Y \in \yspace\) with the target distribution. This change of variables helps clarify that if \g{} is a differentiable function of \(\Y{} = \reparam(\Z;\rparams)\) then, by the chain rule of derivatives \(\small{\frac{d\g}{d\rparams}} = \small{\frac{d\g}{d\reparam}\frac{d\reparam}{d\rparams}}\), i.e., we can use gradient information to optimize the target distribution's parameters \rparams{}. We now explore the importance of this fact for BO and, in particular, for parallel selection.

\section{Reparameterizing acquisition functions}
\label{sect:reparameterizing}

\begin{table}[t!]
\begin{center}
\scalebox{0.98}{
\begin{tabular}{ |c||c|c|c| }
	\hline
	\multicolumn{4}{|c|}{Examples of reparameterizable acquisition functions}\\
	\hline
	Acquisition function & Parameters & Integrand \g{} & Reparameterization\\
	\hline
		\small{Expected Improvement (EI)}
			& \(\means, \Cov; \alpha\)
			& \(\max\left(0, \max\left(\Y\right) - \alpha\right)\)
			& \(\max\left(0, \max\left(\means + \Chol\Z\right) - \alpha\right)\)\\
		\hline
		\small{Probability of Improvement (PI)}
			& \(\means, \Cov; \alpha, \temperature\)
			& \(\heaviside^{-}\left(\max\left(\Y\right) - \alpha\right)\)
			& \(\sigmoid\left(\frac{\max\left(\means + \Chol\Z\right) - \alpha}{\temperature}\right)\)\\
		\hline
		\small{Upper Confidence Bound (UCB)}
			& \(\means, \Cov; \beta\)
			& \(\max\left(\means + \abs{\alt{\Y} - \means}\right)\)
			& \(\max\left(\means + \sqrt{\nicefrac{\beta\pi}{2}}\abs{\Chol\Z}\right)\)\\
		\hline
		\small{Simple Regret (SR)}
			& \(\means, \Cov\)
			& \(\max\left(\Y\right)\)
			& \(\max\left(\means + \Chol\Z\right)\)\\
		\hline
		\small{Entropy Search (ES)}
			& \(\means, \Cov; \temperature\)
			& \(\heaviside^{+}\left(\Y - \max\left(\Y\right)\right)\)
			& \(\softmax\left(\frac{\means+\Chol\Z}{\temperature}\right)\)\\
		\hline
\end{tabular}
}
\vspace{2pt}
\captionsetup{width=.98\textwidth}
\caption{Above, we use the following notation: Cholesky factor \(\Chol\transpose{\Chol} \triangleq \Cov\); \(\heaviside^{+/-}\) denotes the right-/left-continuous Heaviside step function; \sigmoid{} the sigmoid nonlinearity; \(\alpha\) the improvement threshold; \(\tau\) the temperature parameter described in Section \ref{sect:reparameterizing}; and, random variables \(\alt{\Y} \sim \gaussian\left(\means, \nicefrac{\beta\pi}{2}\Cov\right)\). For Entropy Search, a non-myopic acquisition function, only the innermost integrand (used to approximate \(p_{max}\)) and its corresponding reparameterization are shown.
}
\label{table:reparameterizations}
\end{center}
\vspace{-0.25cm}
\end{table}

As is arguably the natural way of expressing uncertainty over interrelated values, beliefs \(p(\Y|\X)\) over the \poolsize{} outcomes for pool \X{} are typically defined in terms of a multivariate normal distribution \(\gaussian(\means, \Cov)\). In the context of the reparameterization trick, the corresponding deterministic mapping for Gaussian random variables \Y{} is \(\reparam(\Z; \means, \Cov) \triangleq \means + \Chol\Z\), where \Chol{} denotes the Cholesky factor of \Cov{}, \st{} \(\Chol\transpose{\Chol} = \Cov\) and \(\Z \sim \gaussian(\zeros, \identity)\). Rewriting \eqref{eq:acquisition_integral} as a Gaussian integral and reparameterizing, we have
\begin{align}
\G(\X; \gparams) 
\label{eq:mvn_integral_reparam}
	&=\int_{\lbounds}^{\ubounds}
		\g(\Y; \gparams)
		\gaussian(\Y; \means, \Cov)
		d\Y
	= \int_{\transform{\lbounds}}^{\transform{\ubounds}}
		\g(\means + \Chol\Z; \transform{\gparams})
		\gaussian(\Z; \zeros, \identity)
		d\Z\,,
\end{align}
where each of the \poolsize{} terms \(\transform{c}_i\) in both \(\transform{\lbounds}\) and \(\transform{\ubounds}\) is transformed as \(\transform{c}_{i} = (c_i - \mean_{i} - \sum_{j<i}\chol_{i j}\z_{j})/\chol_{ii}\) and where values in \transform{\gparams} have similarly been mapped to \zspace{}. By taking the gradient of \(\G(\X; \gparams)\) \wrt{} model-based posterior \(\gaussian(\means, \Cov) = \model(\X)\)  and further differentiating through the model to inputs \X{}, we can perform gradient ascent on acquisition values.\footnote{Parameters associated with model \model{} are not differentiated through and are therefore omitted for clarity.} 

\begin{figure}[t]
\begin{center}
\includegraphics[width=\linewidth]{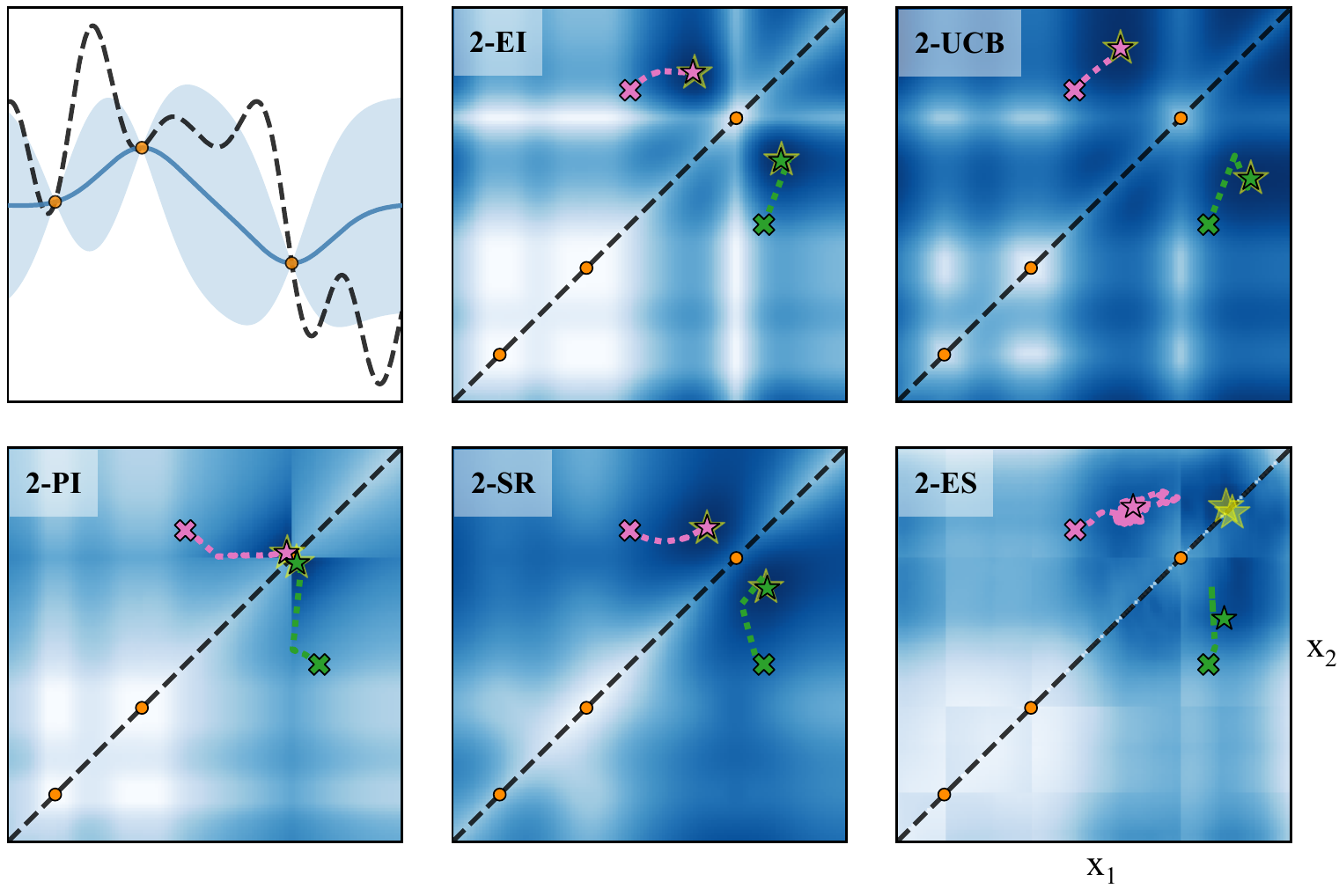}
\vspace{-20pt}
\caption{Top left: GP-based posterior over 1-dimensional black-box \f{} given three initial observations (orange dots). Remaining: Response surfaces of various acquisition functions for pool-size \(\poolsize = 2\). From `\(\boldsymbol{\times}\)' to `\(\bigwhitestar\)', paths explored by gradient descent (green) and stochastic gradient descent (pink) when optimizing the various acquisition functions. Dashed horizontal lines denote axes of symmetry and large `\(\bigwhitestar\)' (yellow) indicate the global maximum of each acquisition function.}
\label{fig:aquisition_surfaces}
\end{center}
\end{figure}

When Monte Carlo integrating \eqref{eq:mvn_integral_reparam}, an unbiased estimate to the acquisition gradients is then
\begin{align}
\label{eq:reparam_acquisition_grad}
\frac{d\G(\X; \gparams)}{d\X} 
	\approx 
		\frac{1}{n}\sum\nolimits_{k=1}^{n}
		\frac{d\g(\Y_{k}; \gparams)}{d\Y_{k}}
		\frac{d\Y_{k}}{d\model(\X)}
		\frac{d\model(\X)}{d\X},
\end{align}
where, by minor abuse of notation, we have substituted in \(\Y_{k} = \reparam(\Z_{k}; \model(\X))\). The availability of gradient information is especially important for \(\poolsize > 1\), both because parallel acquisition functions are generally intractable and because the dimensionality of the acquisition space scales linearly in \(\poolsize\).

Examples of well-known acquisition functions amenable to this treatment are presented in Table~\ref{table:reparameterizations}. Figure~\ref{fig:aquisition_surfaces} provides a visual example of the corresponding (stochastic) gradient ascent process, for each of the five acquisition functions shown in the table. Before going further, several points of interest in Table~\ref{table:reparameterizations} warrant attention:
\begin{enumerate}[leftmargin=14pt,topsep=0pt,itemsep=1ex,parsep=1ex]
\item{\textbf{Parallelizing UCB}: To the best of our knowledge, the integral representation of \UCB{} is novel and leads to the first truly parallel formulation of \UCB{} (\qUCB). Relevantly, using the reparameterization trick greatly simplifies the associated derivation. As with other acquisition functions discussed here, \qUCB{} can be efficiently estimated via Monte Carlo and optimized using gradients. For the complete derivation and related formulae, please refer to Appendix~\ref{sect:q_ucb}.}
\item{\textbf{Relaxing Heaviside step functions}: Both Probability of Improvement (\PI{}) and Entropy Search (\ES{}) contain Heaviside step functions, whose derivatives are Dirac delta functions. Since these gradients are zero a.e., we instead propose the use of a \softmax{} function with temperature parameter \temperature{}. This combination has the appealing property that the resulting approximation becomes exact as \(\temperature \to 0\), a property recently exploited in~\cite{jang2016categorical,maddison2016concrete}. To the extent that this soft approximation introduces an additional source of error, we argue that this downside is largely outweighed by the availability of informative gradients, which enable us to greatly reduce optimization error~\cite{bousquet2008}.}
\item{\textbf{Differentiating though the max()}: Many acquisition functions, such as \EI{}, use the max operator. While not technically differentiable, this operator is known to be subdifferentiable and affords well-behaved (sub)gradients.
}
\end{enumerate}

\section{Experiments}
\label{sect:experiments}

\begin{figure}
\begin{center}
\includegraphics[width=\linewidth]{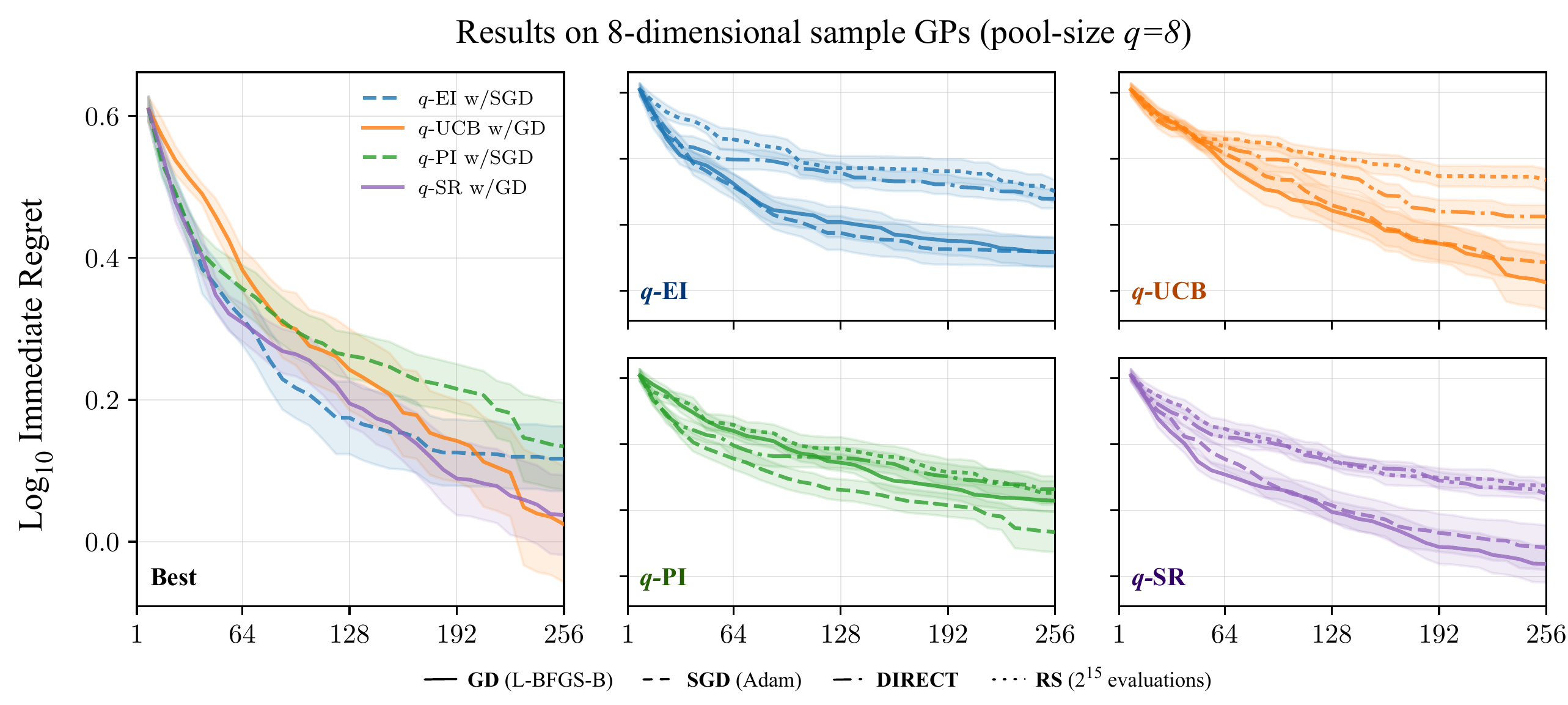}
\vspace{-18pt}
\caption{Left: For equivalent runtimes, best average case performance of each acquisition function given 256 evaluations of 8-dimensional samples from a GP prior with known hyperparameters when choosing pool-size \(\poolsize=8\) queries in parallel. Remaining: Performance of individual acquisition functions for different optimizers thereof.}
\label{fig:log_regret}
\end{center}
\end{figure}

As baselines, we compared gradient-based approaches to optimizing acquisition functions with Random Search \cite{bergstra12} and Dividing Rectangles \cite{jones1993lipschitzian} based ones.
For stochastic gradient descent (SGD), we experimented with several off-the-shelf optimizers; of these, \adam{} \cite{kingma2014adam} produced the best results and is reported here. Similarly, we tested various batch-sizes \(\batchsize\) and report results for \(\batchsize=64\). For gradient descent (GD), we used a standard implementation of L-BFGS-B \cite{zhu1997algorithm}. In both cases, gradient-based optimizers were run using 32 starting points sampled from the acquisition function. Finally, for both \qPI{} and \qES{}, we set the temperature \(\temperature{} = 0.01\); and, for \qUCB{}, we set the confidence parameter \(\beta = \scriptstyle{\sqrt{3}}\).

Prior to running our experiments, we configured each acquisition function optimizer such that its runtime approximately matched that of the others. Further details regarding our experiments, including individual runtimes, are provided in Appendix~\ref{sect:experiment_details}.

To help reduce the number of potentially confounding variables, we experimented on 8-dimensional tasks drawn from a Gaussian process prior with known hyperparameters. For each combination of acquisition function and optimizer, trials began with \poolsize{} randomly chosen observations and iterated by choosing \poolsize{} queries at a time.\footnote{Methods discussed here extend to the parallel asynchronous setting; but, we did not explore this option.} Each pair was run on a total of 16 sampled tasks, with results shown in Figure~\ref{fig:log_regret}. Across acquisition functions, gradient-based strategies markedly outperformed gradient-free alternatives. Further, stochastic and deterministic gradient methods delivered comparable performance.

\section{Conclusion}
\label{sect:conclusion}
We show how many popular acquisition functions can be written as Gaussian integrals amenable to the reparameterization trick. By reparameterizing these integrals, we clarify the differentiability of their Monte Carlo estimates and, in turn, provide a generalized method for using gradients to optimize acquisition values. Our results clearly demonstrate the superiority of gradient-based approaches for optimizing acquisition functions, even in modest dimensional cases. Further, we show how, by looking at the associated integrals through the lens of the reparameterization trick, the difficult process of deriving theoretically sound acquisition functions may be greatly simplified.

\newpage
\section*{Acknowledgments}
The support of the EPSRC Centre for Doctoral Training in High Performance Embedded and Distributed Systems (reference EP/L016796/1) is gratefully acknowledged. This work has partly been supported by the European Research Council (ERC) under the European Union's Horizon 2020 research and innovation programme under grant no. 716721.

\footnotesize
\linespread{1.0}\selectfont
\setlength{\bibsep}{5pt}
\bibliography{references}

\newpage
\normalsize
\appendix
\section{Parallel Upper Confidence Bound (\qUCB{})}
\label{sect:q_ucb}
Working backward through~\eqref{eq:mvn_integral_reparam}, we derive an exact expression for parallel \UCB{}. In doing so, we begin with the definition
\begin{align}
\label{eq:integral_def}
\int_{0}^{\infty} \sqrt{2\pi}\y \gaussian(\y; 0, \var)d\y = \frac{1}{2}\int_{-\infty}^{\infty}\abs{\sqrt{2\pi}\stdv\z} \gaussian(\z; 0, 1)d\z = \stdv,
\end{align}
where \(\abs{\cdot}\) denotes the (element-wise) absolute value operator.\footnote{This definition comes directly from the standard integral identity \(\int_{0}^{\infty}xe^{-ax^2}dx = \nicefrac{1}{2a}\).} Using this fact and given \(\z \sim \gaussian(0, 1)\), let \(\alt{\stdv}^{2} \triangleq (\nicefrac{\beta\pi}{2})\var\) such that \(\expectation{\abs{\alt{\stdv}\z}} = \beta^{\nicefrac{1}{2}}\stdv\). Under this notation, marginal \UCB{} can be expressed as
\begin{align}
\label{eq:mUCB_traditional}
\mUCB(\x; \beta)
	&=\mean + \beta^{\nicefrac{1}{2}}\stdv\\
\label{eq:mUCB_reparam}
	&=\int_{-\infty}^{\infty}\mean + \abs{\alt{\stdv} \z} \gaussian(\z; 0, 1) d\z\\
\label{eq:mUCB_y}
    &= \int_{-\infty}^{\infty} \mean + \abs{\y - \mean}\gaussian(\y; \mean, \alt{\stdv}^2)d\y
\end{align}
where \((\mean, \var)\) parameterize a Gaussian posterior over \(\y = f(\x{})\). This integral form of \mUCB{} is advantageous precisely because it naturally lends itself to the generalized expression
\begin{align}
\label{eq:qUCB}
\qUCB(\X; \beta)
	&= \int_{\boldsymbol{-\infty}}^{\boldsymbol{\infty}}
			\max(\means + \abs{\Y - \means})
			\gaussian(\Y; \means, \alt{\Cov})
			d\Y\\
\label{eq:qUCB_reparam}
	&= \int_{\boldsymbol{-\infty}}^{\boldsymbol{\infty}}
			\max(\means + \abs{\alt{\Chol}\Z}) 
			\gaussian(\Z; \zeros, \identity)
			d\Z\\
\label{eq:qUCB_mc}
	&\approx 
    	\frac{1}{n} \sum\nolimits^{\nsamples}_{k=1} \max(\means + \abs{\alt{\Chol}\Z_{k}})\quad \text{ for~~~} \Z_{k} \sim \gaussian(\zeros, \identity)\,,
\end{align}
where \(\alt{\Chol}\transpose{\alt{\Chol}} = \alt{\Cov} \triangleq (\nicefrac{\beta\pi}{2})\Cov\). This representation has the requisite property that, for any size \(\transform{\poolsize} \le \poolsize\) subset of \X, the value obtained when marginalizing out the remaining \(\poolsize - \transform{\poolsize}\) terms is its \transform{\poolsize}-UCB value.

Previous methods for parallelizing \UCB{} have approached the problem by imitating a purely sequential strategy \cite{contal2013parallel,desautels2014parallelizing}. Because a fully Bayesian approach to sequential selection generally involves an exponential number of posteriors, these works incorporate various well-chosen heuristics for the purpose of efficiently approximate parallel \UCB{}.\footnote{Due to the stochastic nature of the mean updates, the number of posteriors grows exponentially in \poolsize.}. By directly addressing the associated \(\poolsize{}\)-dimensional integral however, Equation~\eqref{eq:qUCB_mc} avoids the need for such approximations and, instead, unbiasedly estimates the true value.

Finally, the special case of marginal \UCB~\eqref{eq:mUCB_reparam} can be further simplified as
\begin{align}
\label{eq:mUCB_simple}
\mUCB(\x; \beta)
    =\mean + 2\int_{0}^{\infty}\alt{\stdv} \z \gaussian(\z; 0, 1)dz
    =\int_{\mu}^{\infty} \y \gaussian(\y; \mu, 2\pi\beta\var) d\y,
\end{align}
revealing an intuitive form, namely, the expectation of a Gaussian random variable (with rescaled covariance) above its mean.

\newpage
\section{Experiment details}
\label{sect:experiment_details}

\begin{table}[h]
\begin{center}
\vspace{-4pt}
\begin{tabular}{ |c||c|c|c|c| }
	\hline
	\multicolumn{5}{|c|}{Runtimes of acquisition function optimizers}\\
	\hline
	Optimizer & \qEI & \qUCB & \qPI & \qSR\\
	\hline 
    Random Search (RS) & \(23.9 \pm 2.3\) & \(17.8 \pm 1.6\) & \(20.1 \pm 1.9\) & \(20.4 \pm 1.9\)\\
	\hline
    Dividing Rectangles (\direct{}) & \(19.8 \pm 1.5\) & \(21.5 \pm 1.9\) & \(21.0 \pm 1.7\) & \(20.2 \pm 1.5\)\\
    \hline
	GD (L-BFGS-B) & \(19.9 \pm 9.0\) & \(18.2 \pm 1.4\) & \(17.6 \pm 7.8\) & \(13.7 \pm 1.2\)\\
	\hline
	SGD (\adam{}) & \(17.6 \pm 9.2\) & \(13.6 \pm 5.8\) & \(15.6 \pm 6.0\) & \(15.4 \pm 5.9\)\\
	\hline
\end{tabular}
\vspace{2pt}
\captionsetup{width=.89\textwidth}
\caption{Average runtime in seconds for each combination of acquisition function and optimizer when choosing the next pool of inputs. Reported numbers denote the mean and standard deviation of recorded wall-clock times.}
\label{table:runtimes}
\end{center}
\vspace{-10pt}
\end{table}

To provide fair comparison between acquisition function optimizers, efforts were made to approximately match their respective runtimes. First, Random Search was run using a set of \(2^{15}\) uniform random pools \X{}, at each step during BO. Subsequently, RS's average runtime, measured over a handful of preliminary trials, was used as a target value when configuring the remaining optimizers. Table~\ref{table:runtimes} provides individual runtimes for each combination of acquisition function and optimizer.

For stochastic gradient descent, we tested the following optimizers: SGD with momentum, \rmsprop{}, and \adam{}. Trials were run using batch-sizes \(\batchsize \in \{32, 64, 128, 256\}\), each time tuning the number of SGD steps for equivalent runtimes. Of the tested configurations, \(1024\) steps using \(\batchsize=64\) delivered the best performance.

\end{document}